\documentclass[10pt, a4paper]{article}
\usepackage{lrec2022} 
\usepackage{multibib}
\newcites{languageresource}{Language Resources}
\usepackage{graphicx}
\usepackage{tabularx,booktabs}
\usepackage{soul}
\usepackage{multirow}
\usepackage{titlesec}
\titleformat{\section}{\normalfont\large\bfseries\center}{\thesection.}{1em}{}
\titleformat{\subsection}{\normalfont\SmallTitleFont\bfseries\raggedright}{\thesubsection.}{1em}{}
\titleformat{\subsubsection}{\normalfont\normalsize\bfseries\raggedright}{\thesubsubsection.}{1em}{}
\renewcommand\thesection{\arabic{section}}
\renewcommand\thesubsection{\thesection.\arabic{subsection}}
\renewcommand\thesubsubsection{\thesubsection.\arabic{subsubsection}}

\usepackage{epstopdf}
\usepackage[utf8]{inputenc}

\usepackage{hyperref}
\usepackage{xstring}

\usepackage{color}
\usepackage{euflag}

\title{What a Creole Wants, What a Creole Needs}

\name{Heather Lent$^{1}$, Kelechi Ogueji$^{2}$, Miryam de Lhoneux$^{1,3,4}$, Orevaoghene Ahia$^{5}$, Anders Søgaard$^{1}$} 
\address{ 
         $^{1}$ University of Copenhagen, Denmark\\
         $^{2}$ University of Waterloo, Canada\\
         $^{3}$ Uppsala University, Sweden \\
         $^{4}$ KU Leuven, Belgium\\
         $^{5}$ University of Washington, United States \\
         \{hcl, ml, soegaard\}@di.ku.dk\\
         kelechi.ogueji@uwaterloo.ca, oahia@cs.washington.edu}

\abstract{
In recent years, the natural language processing (NLP) community has given increased attention to the disparity of efforts directed towards high-resource languages over low-resource ones. Efforts to remedy this delta often begin with translations of existing English datasets into other languages. However, this approach ignores that different language communities have different needs. We consider a group of low-resource languages, \textit{Creole} languages. Creoles are both largely absent from the NLP literature, and also often ignored by society at large due to stigma, despite these languages having sizable and vibrant communities. We demonstrate, through conversations with Creole experts and surveys of Creole-speaking communities, how the things needed from language technology can change dramatically from one language to another, even when the languages are considered to be very similar to each other, as with Creoles. We discuss the prominent themes arising from these conversations, and ultimately demonstrate that useful language technology cannot be built without involving the relevant community.
 \\ \newline  }

\begin{document}

\maketitleabstract

\section{Introduction}
The field of natural language processing (NLP) has become aware that most of the world's languages are unfortunately under-represented, or entirely absent, from the field's body of work \cite{Joshi2020TheSA}.
In recent years, there has been a push in efforts to ameliorate this discrepancy \cite{nekoto-etal-2020-participatory,mirzakhalov-etal-2021-large,ogueji-etal-2021-small}.  
Among these low-resourced languages\footnote{This term is often largely ambiguous, and all ``low-resource" languages should not be conflated together into one large group, but rather considered independently, in the context of its speakers, their culture, and their needs.} are \textit{Creole} languages, which are particularly under-resourced due to barriers like societal stigma \cite{siegel1999stigmatized}, despite the fact that these languages are spoken by many people globally. 
One line of work has focused on creating datasets for low-resource languages
via the translation of existing high-resource language datasets \cite{conneau-etal-2018-xnli,artetxe-etal-2020-cross,budur-etal-2020-data}. 
Despite the popularity of this method, it poses several issues, which can negatively affect the communities of these low-resource languages.
One such issue lies in translation artifacts, which have been shown to have notable impacts on the performance of models trained with such datasets \cite{artetxe-etal-2020-translation}.
Furthermore, translated datasets are often simplified and unnatural, a phenomenon referred to as \textit{translationese} \cite{volansky-translationese-2013}.
This has also been shown to adversely affect the evaluation of machine translation models \cite{graham-etal-2020-statistical}.
Creoles, too, are not immune to the shortcomings of this approach. Moreover, many translated datasets will simply not be relevant to communities speaking a Creole language, as concepts relevant to the original high-resource source language are subsequently translated into the low-resource language, despite being irrelevant to people or cultures speaking the language \cite{Liu2021VisuallyGR}. 
For example, sentences about American football or the American Thanksgiving holiday are simply not relevant or necessary for speakers of Creole languages. The same mismatch also applies to other more geographical-specific domain information present in the data, such as landmarks or landscapes.
All of these show that, while there may be good intentions behind this approach, it could potentially lead to poor models for speakers of low-resource languages and even to the creation of tools of little use or relevance for Creole speakers. 

Meanwhile, works such as \newcite{hu-etal-2011-value} concretely demonstrate how crowd-sourcing data from target-language speakers, even if monolingual, leads to improved results for statistical machine translation systems.
While these findings are not up to date with contemporary neural machine translation, involving native speakers minimizes the risk of having non-relevant examples included in the dataset. However, as the authors also note, there can be considerable logistical difficulties of finding native speakers to contribute, even when offering payment. And even if one manages to recruit paid speakers, a large problem still remains: the underlying exploitative nature of treating language speaking communities like data resources to be mined. \newcite{bird-2020-decolonising} discusses in detail these foundational problems within the language technology community, and how, in order to break the cycle of harmful colonialism in our science, we must fundamentally change the relationship between researchers and the language-speaking communities. 
But the only way we can learn this, claims \newcite{bird-2020-decolonising}, is by establishing a respectful, ``feedback/collaboration loop", and necessarily involving community members in our research.

Following the work of \newcite{bird-2020-decolonising}, in this work, we focus on the problem of creating resources for low-resource languages, in this case Creoles, and the inherent presupposition by researchers of what technologies are indeed wanted and needed by the communities speaking those languages. While many researchers may assume that the ``best-case scenario" for all languages would be to have all language technologies \textit{equally} available, the fact of the matter is that many communities have very specific wants and needs of language technology, as well as language technologies that are notably unwelcome, even though they are a commonplace for high-resource languages. Disregarding the needs of a language community can lead to misuse of finite resources on creating unnecessary datasets or technologies while leaving the community's highest priorities neglected. And finally, when researchers assume what technologies are wanted on the behalf of a community, it inherently alienates that community and takes away their agency \cite{bird-2020-decolonising}. In this work, we explore how the needs of different Creole-speaking communities vary wildly from one another, and we demonstrate the need to establish respectful relationships with experts and communities, in order to make truly useful language technology. 

Our contributions in this work are as follows: 
\begin{itemize}
    \item We present a survey of Creole NLP, and discussion of features from Creole languages that present unique challenges to existing NLP workflows.
    \item We discuss important considerations, gleaned from conversations with experts, and a survey of Creole language speakers.
    \item We propose a Creole continuum for language technology, as a guiding framework of research considerations, to help NLP researchers planning to work on Creoles.
\end{itemize}

\section{Background}
\label{background}
Today, Creole languages are spoken widely throughout the Caribbean and West Africa, as well as parts of South America, Asia, Australia and the Pacific. 
Creoles have long captured the attention of linguists due to their unique, and sometimes tragic\footnote{For example, Caribbean Creoles resulted from the displacement of African peoples in the Atlantic slave trade.}, histories with regards to language evolution.
Typically, Creole languages originate from situations in which multiple different languages have come into close contact with each other \cite{thomason1992language}. 
The exact process of how a Creole language is ``born" (i.e. Creole genesis), as well as discussion of which linguistic features a Creole inherited from the various ``parent" languages, have been the subject of intense and ongoing linguistic debate for decades \cite{mervyn1971acculturation,bickerton1984language,muysken1986substrata,Sessarego2020NotAG}. On one hand, some believe that Creoles themselves form a unique typological class of languages, with a separate place on the phylogenetic tree of languages (i.e. Creole exceptionalism, \newcite{bickerton1984language}). Linguists supporting Creole exceptionalism typically claim that Creoles are more simple than other languages \cite{parkvall2008simplicity}, for example, lacking in complex morphology \cite{mcwhorter1998identifying}. On the other hand, others argue that there are no grounds to claim that Creole evolution is especially different from the language evolution of so-called ``normal" \cite{degraff2003against,degraff2005linguists}. And indeed, Creoles do exhibit behaviors just as complex as non-Creole languages \cite{degraff2001origin}, including complex morphology \cite{henriderivation}.

Moreover, some criticisms of Creole exceptionalism also examine how the history of Creole studies itself has unfortunately been riddled with discrimination and racism \cite{degraff2005linguists}. 
In the past, Creoles were often considered to be something short of a full-fledged language (or, more harshly, ``degenerate variants or dialects of their parent languages"\footnote{\url{https://en.wikipedia.org/wiki/Creole_language\#Overview}}). 
According to \newcite{kouwenberg2009handbook}, ``A part of the legacy of slavery in the Caribbean and elsewhere has been the stigmatization of the languages associated with slaves ... [the] willingness to apply the concept of linguistic relativism -- whereby every language is understood to be complete and valid -- may have been extended to Hopi and Hausa\footnote{Hopi is an Native American indigenous language from Arizona, United States; Hausa is a Chadic language, spoken in West and Central Africa.}, but it generally stopped short of being extended to Creoles." In line with this, in this work, we hope to raise awareness in the NLP community about why Creoles are important to work with. Beyond being the subject of vibrant linguistic debate, Creoles are often ignored when it comes to language technology, which puts speakers of already often stigmatized languages at a further disadvantage. For the remainder of this section, we will present a survey of existing Creole datasets, a summary of works published on NLP for Creoles, and finally end this section with a discussion of some specific features of Creole languages that are notable within the context of NLP. 

\subsection{Creole Data and Creole NLP}
In this section we will detail existing resources and datasets for Creole languages (including those which are now seemingly defunct), as well as discuss related works actively focused on NLP for Creoles. 

\paragraph{Verified Resources}
Although Creole languages are in general very low-resourced, the datasets that do exist vary widely from task to task, as well as from language to language.  \newcite{hagemeijer-etal-2014-gulf} presents an extensive overview of Creole data resources through 2014 for a wide variety of Creoles, many of which are more traditional corpora, (e.g., transcriptions of conversations made by linguists with formal training, or scans of documents originally written in the Creole language); though these may not have the relevant annotations for common NLP tasks. 
\newcite{lent-etal-2021-language} also provides a thorough overview of existing NLP datasets for Haitian Kreyol, Singaporean Colloquial English (Singlish), and Nigerian Pidgin English. In this work, we set about the task of manually verifying each dataset presented by \newcite{hagemeijer-etal-2014-gulf} and \newcite{lent-etal-2021-language}, as well as searching for additional resources. We present all ``verified" datasets in Table \ref{tab:available-data}. Here, we use ``verified" to mean that we could easily find the resource described in the paper, through either a provided URL in a publication, or through a search engine. 

Readers should note that we excluded both extinct Creoles and ostensibly historical Creole data from Table \ref{tab:available-data}. Those interested can see that there are available data for the extinct Virgin Islands Dutch Creole.\footnote{\url{doecreoltaal.com}}
Other historical Creole data include the Corpus of Mauritian Creole Texts \cite{baker2007making}, a collection of texts spanning the 1730 to 1930, and the Surinam Creole Archive (\url{suca.ruhosting.nl}), which should have historical texts for both Sranan Tongo and Saramaccan, although the hyperlinks are presently broken in this website. 

Lastly, to utilize linguistic information about Creoles, the Atlas of Pidgin and Creole Language Structures (APiCS) is an indispensable resource \cite{apics}. APiCS is an extension of the popular WALS resource \cite{wals}, but is solely dedicated to pidgins and Creoles.

\begin{table*}[t]
\centering
\begin{tabular}{lp{0.4\textwidth}p{0.2\textwidth}p{0.15\textwidth}}
\toprule 
\textbf{Language} & \textbf{Resource} & \textbf{Description} & \textbf{Status} \\ \midrule

\multirow{5}{*}{Haitian Kreyol} & Haitian Disaster Response Corpus\ \cite{Munro10crowdsourcedtranslation,WMT:2011}& SMS & {Verified; E-mail authors for access.} \\ \cline{2-4} 
 & CMU Haitian Corpus \url{http://www.speech.cs.cmu.edu/haitian/} & Speech and Text Corpora & {Verified; E-mail authors for access.} \\ \midrule
 
\multirow{2}{*}{Hawaiian Pidgin} & Multilingual Hawai'i Linguistic Landscape Corpus \cite{purschke2021crowdscapes} & Image Repository with Annotations & Verified \\ \midrule
 
\multirow{14}{*}{Nigerian Pidgin} & NaijaSynCor \cite{Bigi2017DevelopingRF} & Speech Recognition & {Verified} \\ \cline{2-4} 
 & JW300 Corpus \cite{agic-vulic-2019-jw300} & Parallel Texts for Machine Translation & {Verified} \\ \cline{2-4}
 & Pidgin UNMT \cite{Ogueji2019PidginUNMTUN} & Monolingual Texts for Machine Translation & {Verified} \\ \cline{2-4}
& Naija-English Codeswitching Corpus \cite{obighosh2019naija} &  News Articles with Comments; Annotated for code switching & {Verified} \\ \cline{2-4} 
& Surface-Syntactic UD Treebank for Naija \cite{caron-etal-2019-surface} & Universal Dependencies & {Verified} \\ \cline{2-4} 
 & Speech-to-Text Nigerian Pidgin Dataset \cite{Ajisafe2020TowardsET} & Speech Recognition & {Verified} \\ \cline{2-4} 
 & NaijaNER \cite{Oyewusi2021NaijaNERC} & Named Entity Recognition & {Verified} \\ \cline{2-4}
 & Masakhaner \cite{adelani-etal-2021-masakhaner} & Named Entity Recognition & {Verified} \\ \cline{2-4} 
  & NaijaSenti \cite{https://doi.org/10.48550/arxiv.2201.08277} & Sentiment Analysis & {Verified} \\ \midrule

Reunionese Creole \& & Creolica & Text and Short Stories & {Verified} \\
Seychellois Creole & \url{http://creolica.net/} & in HTML or PDFs & \\\midrule

\multirow{5}{*}{Singlish} & National University of Singapore SMS Corpus \cite{10635_137343} & SMS & {Verified} \\ \cline{2-4}
& Universal Dependencies for Colloquial Singaporean English \cite{wang-etal-2017-universal} & UD Treebank & {Verified} \\ \cline{2-4}
& Webcrawler for Singaporean Hardware Forum \cite{tan-etal-2020-mind} & Webcrawler & {Verified} \\ \midrule

{Sri Lankan Malay} & The Language Archive \citelanguageresource{dobes} & Audio and XML & {Verified}\\
(Endangered) & & & \\ \midrule

\end{tabular}
\caption{Descriptions of every Creole resource or dataset that we could identify and also verify as being readily available online.}
\label{tab:available-data}
\end{table*}

\paragraph{Unverified Resources}
Unfortunately many of the Creole corpora reviewed in \newcite{hagemeijer-etal-2014-gulf} are no longer available, with broken URLs. We describe any resource as ``not verifiable", when we cannot track down the resource through the combination of a URL, a simple web search, or through the original publication. These resources may still exist, but they are too difficult to find with a reasonable effort made. The list of ``not verifiable" resources can be found in Table \ref{tab:available-undata}. We hope that highlighting the ``not verifiable" datasets can serve as a call to action in the field, to consider long term data hosting solutions.
In order to make the information we gathered about datasets useful in the long-term, we release a community-based webpage.\footnote{\url{https://creole-nlp.github.io/}} It is hosted on github pages and allows pull requests so that community members can help us maintain up-to-date information about data available for Creoles.

Moreover, in this section, we would also like to discuss book-based corpora. We do not include them in Table \ref{tab:available-undata}, as considerable work would need to be done to digitize these datasets, before they can be usable for most NLP tasks. Still, these resources could be useful for those wanting to work with some Creole languages, not listed in Table \ref{tab:available-data} or Table \ref{tab:available-undata}. Creole corpora documented in books include the Corpus of Written British Creole \cite{sebba1998phonology}, a corpus of folktales in Tok Pisin \cite{slone2001one}, and a corpus of Jamaican Creole \cite{hinrichs2006codeswitching}. We also found the following additional resources described by \newcite{kouwenberg2009handbook} :
transcripts of Guyanese Creole \cite{rickford1987dimensions}, transcripts of English-based Central American Creoles were introduced by \cite{holm1982central}, and a corpus of various French-based Creoles, such as Louisiana Creole and Reunionese Creole \cite{corne1999french}. 

\begin{table*}[t]
\centering
\begin{tabular}{lp{0.4\textwidth}p{0.2\textwidth}p{0.15\textwidth}}
\toprule 
\textbf{Language} & \textbf{Resource} & \textbf{Description} & \textbf{Status} \\ \midrule
\multirow{3}{*}{Antillean Creole} & CREOLORAL \url{http://ircom.corpus-ir.fr/site/description_projet.php?projet=CREOLORAL} & Audio, Transcriptions, and Translations & {Not verifiable} \\ \midrule

Bastimentos Creole & Endangered Language Archive  & Audio, Video, & Not verifiable; \\
English & \url{http://elar.soas.ac.uk/deposit/0171} & Transcriptions, Translations & Membership required \\ \midrule

Gulf of Guinea Creoles & The Gulf of Guinea Creole Corpora \cite{hagemeijer-etal-2014-gulf}  & Document Scans and Transcriptions & Limited Verifiability \\\midrule

\multirow{2}{*}{Haitian Kreyol} & Corpus of Northern Haitian Creole \url{https://www.indiana.edu/~Creole/} & Audio and Transcription & Not verifiable \\ \midrule

Malaccan Portuguese & Endangered Language Archive  & Audio, Video, & Not verifiable; \\
Creole & \url{http://elar.soas.ac.uk/deposit/0123} & Transcriptions, Translations & Membership required \\ \midrule 
 
\multirow{3}{*}{Mauritian Creole} & ALLEX Project \url{http://www.edd.uio.no/allex/corpus/africanlang.html} & Concordance of 200k Words & {Not verifiable} \\\midrule

\multirow{2}{*}{Nigerian Pidgin} & Nigerian Pidgin Tweets \cite{Oyewusi2020SemanticEO} & Sentiment Analysis &  {Not Verifiable}\\ \midrule

\multirow{1}{*}{Portuguese Creole} & CreolData \cite{schang2005creoldata} & Lexical Database & Not verifiable \\ \midrule

\multirow{3}{*}{Singlish} & Singlish Sentiment Lexicon \cite{DBLP:journals/corr/BajpaiPHC17} & Knowledge Base  & Not Verifiable \\ \cline{2-4}
& Singlish SenticNet \cite{8628796} & Sentiment Resource & Not Verifiable \\ \midrule

\end{tabular}
\caption{Description of Creole datasets presented in our resource survey, which we were not able to verify the existence of. Note here that ``Gulf of Guinea Creoles" refers to a collection of four distinct Creole languages: Santome, Angolar, Principense, and Fa d'Ambo.}
\label{tab:available-undata}
\end{table*}

\paragraph{NLP for Creoles}
Creole languages, though largely absent from the NLP literature, have been investigated directly in a small number of works. Of the few works actively focused on Creoles, two works explore directly Creole genesis in the context of computational linguistics.
First, \newcite{DavalMarkussen2012ExplorationsIC} employ phylogenetic tools to explore whether Creole langauges form a unique typological group. By treating each Creole as a list of binary linguistic features, including data from WALS \cite{wals}, they analyze the output of a phylogenetic network program \cite{huson2006application}, to inform their investigation. The overall conclusion made by \newcite{DavalMarkussen2012ExplorationsIC}, was that Creoles indeed formed their own distinct typological class, distinguishable from non-Creoles.
However, this work was later refuted by \newcite{Murawaki2016StatisticalMO}, who argued that the study by \newcite{DavalMarkussen2012ExplorationsIC} had some methodological shortcomings. 
Notably, \newcite{Murawaki2016StatisticalMO} use APiCS features \cite{apics} to encode Creoles, and utilize different approaches for language evolution modeling, to reach the final conclusion that Creoles are \textit{not} typologically distinct from non-Creole languages. 

Meanwhile, \newcite{lent-etal-2021-language} explored the question of how to effectively build language models for three Creole languages (Haitian Kreyol, Singaporean Colloquial English, and Nigerian Pidgin). Their approach involved experimenting with distributionally robust objectives \cite{oren2019distributionally}, to ascertain whether data from a Creole's ``parent'' languages could help the language model to be more robust. In the end, they found that straightforward training of language models for Creoles, without adding information from their related languages, produced the strongest results, thus highlighting the relative stability of Creoles. 

Finally, there have been a handful of other works aiming to develop NLP algorithms usable for end users, primarily in the area of machine translation, for Creoles like Haitian Kreyol, Mauritian Creole, and Nigeran Pidgin \cite{WMT:2011,dabre-etal-2014-anou,millour-fort-2020-text,Ahia2020TowardsSA}.

\subsection{Notable Features of Creoles}
Many Creole languages are noteworthy for their large capacity for linguistic variation. A speaker's individual style of Creole can vary dramatically depending on social factors, such as their age, ethnicity, geography, and social status. These variations can manifest themselves in different linguistic functions of the Creole, for instance, in the chosen syntax, morphology, or lexical choices \cite{DBLP:journals/corr/BajpaiPHC17}.
Below, we discuss other features of some (not all) Creoles, that are particularly notable in the context for NLP.

\paragraph{Societal Stigma vs Recognized Status}
Creole languages are infamously stigmatised \cite{mervyn1971acculturation,siegel1999stigmatized}. To this day, prejudice against Creole languages has thwarted Creole-based education being made available to Creole speakers, for example. The relative status of a language can change drastically, from Creole to Creole. For instance, use of Singlish has been actively discouraged by government officials, citing the need to ``Speak Good English".\footnote{\url{https://en.wikipedia.org/wiki/Speak_Good_English_Movement}} Meanwhile, a handful of other countries have come to embrace Creole (to varying degrees) in their education system, such as Haitian Kreyol, Papiamento, Seychellois Creole, and Tok Pisin \cite{kouwenberg2009handbook}. 
The relative celebration or suppression a Creole receives will certainly impact who is speaking the Creole language, and how they will use it. 

\paragraph{Spoken Languages} 
Today a large number of Creole languages exist primarily, or almost entirely, as a spoken language only (this can also be a consequence of high stigmatization, as explained in the paragraph above \cite{sebba1997contact}). If Creole speakers are not typically writing in the language, development of text-based NLP methods may be largely superfluous, unless members of that community have expressed a desire to begin writing (more) in Creole. Consequently, speech technologies may be more relevant to a large number of Creole speaking communities. 

\paragraph{Non-standardized Orthography or Grammar} 
Writing conventions for Creoles can vary greatly, from Creole to Creole, and even from speaker to speaker. Given that Creoles arise from a complex process involving several parent languages \cite{Sessarego2020NotAG}, and formal writing education in that Creole is not a guarantee for speakers \cite{siegel1999stigmatized}, there is often no standard way of writing them. On one hand, spellings can depend on an individual and informed by their own oral version of the language \cite{millour-fort-2020-text}.  Moreover, spelling and grammar conventions in Creole can also be affected greatly by a speaker's proficiency in that Creole. For instance, native speakers of Nigerian Pidgin may speak a fluent, fast, and strong variety of the Creole (i.e., less diluted with English), while others speak a weaker 
Creole, learned as a second language, characterized by heavy use of just one ancestral Nigerian language. This kind of variety in many cases, as with Nigerian Pidgin, is considered a very positive aspect of a Creole, as it grants speakers a lot of opportunity for nuanced expression.
Given that contemporary NLP methods are typically not robust to such linguistic variation, it is important not to limit Creole speakers to one register of communication \cite{dogruoz-etal-2021-survey}.  

Meanwhile, some Creole languages are undergoing an ongoing cultural shift, with a push towards standardization, in a manner intended to help cultivate a culture of writing in that Creole. For example, in 2014, a language academy was founded for Haitian Kreyol\footnote{\url{https://en.wikipedia.org/wiki/Akademi_Krey\%C3\%B2l_Ayisyen}},\footnote{\url{http://akademikreyol.net/}}. For those planning to work on text-based Creole applications, it is vital to become attuned to the current writing culture of that Creole's community, and be aware of how speakers are wanting to use their Creole in writing.

\paragraph{Bugs or Features?} 
In summary, many of the features discussed above may be perceived as introducing ``challenges" or difficult ``problems" for NLP to grapple with, as these features are not shared with high-resource languages, like English or Mandarin. However, these so-called ``problems for NLP" are often considered positive features by Creole language speakers themselves. We challenge readers not to think of how they can make Creoles work for NLP, but how NLP can work for Creoles. 

\section{What's Wanted and What's Needed} 
In this section, we will give an overview of the key takeaways from our conversations with experts, as well as the major findings from surveying speakers belonging to various Creole speaking communities. 

\begin{figure*}[t]
    \centering
    \includegraphics[scale=.43]{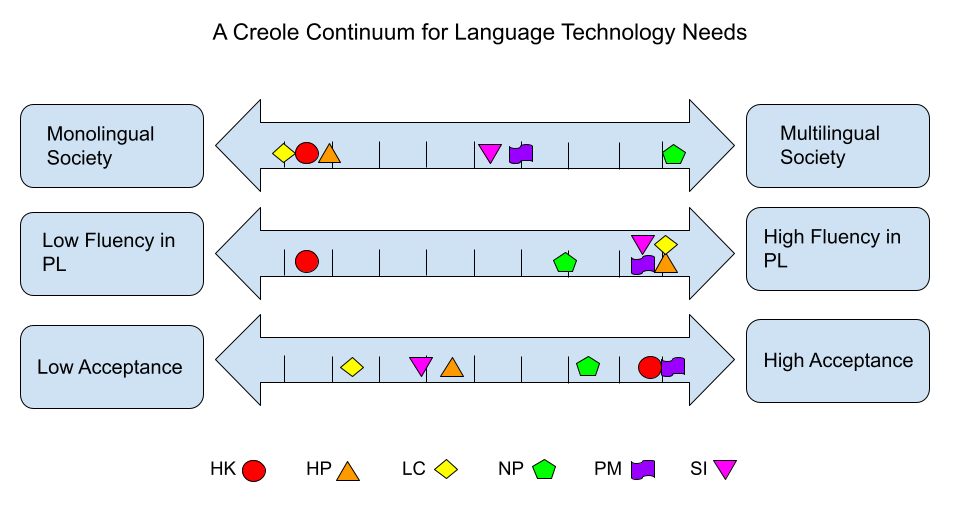}
    \caption{We map a sample of Creole languages to our proposed Creole continuum for language technology. PL here refers to ``Prestige Language". We map Haitian Kreyol (HK), Hawaiian Pidgin (HP), Louisiana Creole (LC), Nigerian Pidgin (NP), Papiamento (PM), Singlish (SI).}
    \label{fig:continuum}
\end{figure*}

\paragraph{Connecting with Experts} As discussed by \newcite{bird-2020-decolonising}, building respectful relationships with the relevant community is absolutely necessary, and reaching out to relevant experts is a great first step towards this direction. For the scope of this work, our definition of an ``expert" is not strict. We consider an expert to be anyone who is engaged in research, education, or other community outreach, somehow involving the Creole. This can include, for example, individuals working at language schools, field linguists doing research in the area, local scientists in any field, or even graduate students who are native speakers of such languages. 
Indeed, there are many reasons to begin by reaching out to experts, before even defining your project. First, despite coming from diverse academic backgrounds, experts across different specialities typically speak the same language of scholarship. Although terminological baggage may still interfere with discussion, generally it is easier for fellow field experts to understand, and empathise with each others' goals, than perhaps others. Moreover, even if the experts are not directly working in your field, they may still be familiar or exposed to it. In establishing this relationship, and learning about each other's research or work, there is also a likelihood that some interests overlap, and the opportunity presents itself that you can also help them, which in turn helps to end the norm of treating low-resource language speakers as resources to extract from, and establish a collaborative relationship \cite{bird-2020-decolonising}. Additionally, experts also have the authority and knowledge to give you an informed ``bird's eye" view of the Creole community, their needs, and desires, as the expert is also a part of it. It's a great (probably necessary) starting point for anyone planning on working on a Creole-language, while not already embedded in the community.

\paragraph {Surveying Creole Speakers} With this in mind, discussion with experts alone runs a large risk of missing out on the thoughts of every day Creole speakers, for whom the language technology is ultimately intended. Thus, their thoughts, opinion, desires, and worries are of utmost importance. For this work, we invited Creole speakers to voice their opinions, and to participate in a survey, through both Twitter and Reddit.
Two points should be noted about this approach: (1) One limitation of this method is that our posts already unfortunately exclude Creole speakers not also speaking English or French, and (2) We attempt to break away from extractive/exploitative research practices by asking only those with additional interest in the topic to fill out the survey (i.e., for those individuals without substantial interest, we try to minimize the time required for them to contribute to the discussion, by asking general, open ended questions). While the best case scenario would have been to compensate people for their time, as \newcite{hu-etal-2011-value} recall, it can be very difficult to find people willing to participate even for payment. Fortunately for this work, we were still able to find a sizeable number of Creole speakers interested in this topic, and willing to have a discussion with us, even if they did not fill out the survey. For the survey, we had 37 participants in total (35 in English, 2 in French), residing in a diverse range of regions (e.g., Caribbean, Africa, North America, Europe, Asia, and the Pacific). We first asked questions about their linguistic background, and use of various languages in daily life. Then, to target NLP wants and needs, we asked more questions about their language use with regards to technology ("e.g. reading/writing SMS on mobile phone, reading/writing on the internet, reading and writing e-mails, interacting with home assistant devices, etc."). For many questions, we included additional prompts, welcoming participants to expand and explain their answers in short-form, which ultimately yielded many important discussion points from the Creole speakers\footnote{Please contact us directly if you would like access to our surveys.}.

For the rest of this section, we will review the consistent themes that arose in our conversations with experts, about the wants and needs of Creole language users. These themes will be further expanded upon by the input provided by Creole speakers from our survey. Again, not all themes will be relevant for every Creole. On the contrary, themes seem to be primarily relevant to Creoles with very specific attributes in common (see Section 4).

\paragraph{Is Language Technology Wanted or Needed?}
As discussed throughout this paper, Creole languages are incredibly diverse, including in the way people want (or don't want) to use these languages to interact with technology. Thus, it should come as little surprise, that the answer to the question: ``Is language technology wanted and/or needed for this language?", can be everything from ``Yes!", ``Some technology would be nice", ``No", and ``Why would you waste time doing that?!", among others.   

Amongst both experts and Creole speaking survey respondents, the answer to this question appeared to be largely contingent on how proficient members of the larger community are in the local, high-prestige language (typically English, French, or Portuguese). For instance, a limited subset of the population of Haiti speaks French, and thus Haitian Kreyol is used in most aspects of every day life, and technology to ease the use of Haitian Kreyol is highly desired. On the other end of the spectrum, experts and speakers of Hawaiian Pidgin had difficulties coming up with reasons why language technology support for their Creole would be particularly useful, as the overwhelming majority of speakers (if not all)  are highly proficient in English.

\paragraph{Current Obstacles} 
In our discussions, some expressed that they already use their Creole for basic tasks, such as texting friends, but that it was not always easy. For example, existing autocomplete or autocorrect software on phones and computers (installed in the relevant high-prestige language, as these technologies are not readily available to Creoles) often automatically ``corrects" Creole spellings or words, and inadvertently suppressing written Creole usage in daily life. 

Another issue that Creole speakers mentioned about existing speech technology, was the lack of support for Creole accents or casual code-switching with commonplace Creole words. For instance, navigational assistants for GPS struggle to understand Hawaiian Pidgin accents, in addition to being unable to pronounce local street names, which can be uttered in Hawaiian Pidgin, but not in Standard American English.
Extending existing speech technology for dominant, high-prestige languages in this space is much desired, and can be preferable over having a separate, Creole-only system. But without these modifications, existing language technology for the local, high-prestige language can actively harm Creole speakers.

\paragraph{Speech Technology}
As discussed in the Background, many Creole languages are used almost exclusively used in spoken conversation. For such Creoles, text-based language technology are likely moot. Although this can change with time, we encourage readers interested in working in Creole spaces to check with experts and communities, to ascertain if text-based technologies are even needed. 

When speech technology was discussed, most Creole languages expressed interest and desire in having speech technology (both text-to-speech and speech-to-text), with the small exception of Creole languages under threat of decreolization (the process by which a Creole ceases to exist), where language revitalization is the dominant concern. But overall, speech technology was perceived by experts and survey respondents to be the most desirable and wanted language technology.

\paragraph{Facilitating Writing}
Some Creole speaking communities already do a lot of writing in their Creole language (despite some obstacles, as we have seen), and/or are trying to foster a culture of writing in the Creole, including standardizing the language. In our conversations with experts and survey responders, we note that there is an expressed need by some Creole communities for basic word processing tools, such as word processors, spell-checkers, grammar-checkers, auto-transcription, etc. However, we found that not all Creole communities welcome \textit{all} of these technologies equally. For example, speakers of Haitian Kreyol mostly welcome spell-checkers, meanwhile speakers of Nigerian Pidgin would eschew these, as it constrains their language use. This point demonstrates how, even when there is a shared desire for a specific kind of language technology, the implementation and specific needs for a Creole can be highly specialized.
Lastly, we note that, just as you must learn to walk before you can run, technologies that ease or improve writing in Creoles may be necessary before Creole speakers could have a need for semantic parsing, for example. 

\paragraph{Question Answering and Machine Translation}
Both question answering (QA) and machine translation (MT) came up as desired technologies for many Creoles, albeit for different reasons.
For Creoles already used online to some extent, QA could improve online search, while MT from Creole into a high resource language, or vice-versa, could provide access to other parts of the world for Creole speakers. Also, MT was cited as desirable for even some endangered Creoles, like Louisiana Creole, as it could help with revitalization. For example, automatic translation from English or French to Louisiana Creole, could allow people to enjoy new domains in Louisiana Creole, and in turn assist with (re)learning the language.

\paragraph{Summary}
Overall, our discussions with Creole experts and every day Creole speakers underscored how diverse the needs of Creoles can be, for even within one group of languages. We hope this discussion, and the themes put forward, can serve as a springboard for those planning to work on NLP for Creoles.

\section{Creole Continuum for Language Technology}
While the previous section demonstrated that Creoles are not a monolith when it comes to wants and needs for language technology, we did observe several patterns, where Creoles seemed to cluster together, based on their language technology needs, depending on a few shared attributes. To this effect, we introduce a Creole continuum for language Technology (inspired by the post-Creole continuum \cite{decamp1971toward}), and propose that there are three key factors that can heavily influence the general needs of a Creole, as follows:
(1) Monolingual, Bilingual, or Multilingual community (in other words, is the Creole a lingua franca, facilitating cross-lingual communication?); and 
(2) General fluency in the relevant prestige language (i.e., do most people also speak the more globally prestigious language, and get on fine, without the Creole?); and
(3) Societal acceptance of Creole (e.g., is the Creole language embraced by society as large, or does the Creole struggle from a bad reputation?).  
We present this continuum in Figure \ref{fig:continuum}, with a small collection of Creole languages, to serve as an example.\footnote{We specifically intend the graph axis to be flexible for interpretation, as different Creoles will have different needs, and strict or concrete axis categories may risk reinforcing existing marginalization.}

The first pattern we would like to draw to the reader's attention to is that the Creoles existing within predominantly monolingual societies, that are also highly fluent in the local prestige language, are those that do not need language technology (Hawaiian Pidgin), at least not beyond revitalization (Louisiana Creole). Also, any time that the larger society exhibits very low fluency in the prestige language, language technology is much more likely to be wanted and needed in these communities (Haitian Kreyol). For other Creoles, it is not so clear cut, though. For example, both Singlish and Papiamento exist in generally multilingual societies, with a majority of speakers also fluent in Dutch and English, respectively, and yet the increased societal acceptance of Papiamento (Papiamento is a recognized language of Aruba), means that speakers are more likely to welcome or express needs for language technologies. Still, Singlish is not to be completely neglected, but due to its lower acceptance, language technology suiting more informal situations (e.g. dialog) will likely be more relevant. And finally, the speakers of languages with high acceptance of Creole, namely Papiamento, Haitian Kreyol, and Nigerian Pidgin, are those who typically have the most clear cut wants and needs from language technology, as they already likely use their Creole to interact with technology.

\section{Conclusion}
In this work, we have demonstrated that Creole languages should be of larger interest to the NLP community, and we provide a survey of resources and NLP research produced for Creoles. In doing this, we have also shown that Creoles cannot be conflated together, if we are to make language technology that is truly useful for a community. Truly, the best approach to developing NLP for Creoles to to get in contact with both experts and community members, and listen earnestly to their wants and needs for language technologies, as well as what is specifically not wanted. 

\section{Acknowledgements}
We would like to thank the Creole language experts and Creole speakers, without whom, this work would not be possible. This includes Michel Degraff (MIT Haiti Initiative), Christina Higgins (Charlene Junko Sato Center for Pidgin, Creole, and Dialect Studies), and our Creole-speaking colleagues in NLP, namely, Samson Tan (National University of Singapore; Salesforce) and Rasul Dent (Université de Lorraine).
Finally, this project has received funding from the European Union’s Horizon 2020 research and innovation programme under the Marie Skłodowska-Curie grant agreement No 801199 (for Heather Lent) \euflag, the Swedish Research Council Grant 2020-00437 (for Miryam de Lhoneux), and the Google Research Award (for
Heather Lent and Anders Søgaard). 

\section{Bibliographical References}\label{reference}

\bibliographystyle{lrec2022-bib}
\bibliography{lrec2022-example}

\section{Language Resource References}
\label{lr:ref}
\bibliographystylelanguageresource{lrec2022-bib}
\bibliographylanguageresource{languageresource}

\end{document}